\documentclass[11pt,a4paper]{article}

\usepackage[affil-it]{authblk}

\usepackage[hyperref]{acl2021}
\usepackage{times}
\usepackage{latexsym}

\usepackage{amsmath,amsfonts,amssymb}
\usepackage{bm}
\usepackage{graphicx,caption}
\usepackage{booktabs}
\usepackage{multicol,multirow}

\usepackage{stfloats}

\DeclareMathOperator*{\argmax}{argmax}

\usepackage{microtype}

\aclfinalcopy 
\def\aclpaperid{1022} 

\title{Dual Slot Selector via Local Reliability Verification \\for Dialogue State Tracking}

\author[1]{\textbf{Jinyu Guo}}
\author[ ]{\textbf{Kai Shuang}\textsuperscript{1}\thanks{\; Corresponding author.}}
\author[1]{\textbf{Jijie Li}}
\author[2]{\textbf{Zihan Wang}}
\affil[1]{State Key Laboratory of Networking and Switching Technology,}
\affil[ ]{Beijing University of Posts and Telecommunications}
\affil[2]{Graduate School of Information Science and Technology, The University of Tokyo}
\affil[ ]{\texttt{\{guojinyu, shuangk, lijijie\}@bupt.edu.cn}}
\affil[ ]{\texttt{zwang@tkl.iis.u-tokyo.ac.jp}}

\date{}

\begin{document}
\maketitle
\begin{abstract}
The goal of dialogue state tracking (DST) is to predict the current dialogue state given all previous dialogue contexts. Existing approaches generally predict the dialogue state at every turn from scratch. However, the overwhelming majority of the slots in each turn should simply inherit the slot values from the previous turn. Therefore, the mechanism of treating slots equally in each turn not only is inefficient but also may lead to additional errors because of the redundant slot value generation. To address this problem, we devise the two-stage DSS-DST which consists of the Dual Slot Selector based on the current turn dialogue, and the Slot Value Generator based on the dialogue history. The Dual Slot Selector determines each slot whether to update slot value or to inherit the slot value from the previous turn from two aspects: (1) if there is a strong relationship between it and the current turn dialogue utterances; (2) if a slot value with high reliability can be obtained for it through the current turn dialogue. The slots selected to be updated are permitted to enter the Slot Value Generator to update values by a hybrid method, while the other slots directly inherit the values from the previous turn. Empirical results show that our method achieves 56.93\%, 60.73\%, and 58.04\% joint accuracy on MultiWOZ 2.0, MultiWOZ 2.1, and MultiWOZ 2.2 datasets respectively and achieves a new state-of-the-art performance with significant improvements. \footnote{Code is available at \\https://github.com/guojinyu88/DSSDST}
\end{abstract}

\section{Introduction}

Task-oriented dialogue has attracted increasing attention in both the research and industry communities. As a key component in task-oriented dialogue systems, Dialogue State Tracking (DST) aims to extract user goals or intents and represent them as a compact dialogue state in the form of slot-value pairs of each turn dialogue. DST is an essential part of dialogue management in task-oriented dialogue systems, where the next dialogue system action is selected based on the current dialogue state.

Early dialogue state tracking approaches extract value for each slot predefined in a single domain~\cite{williams2014dialog,henderson2014second,henderson2014third}. These methods can be directly adapted to multi-domain conversations by replacing slots in a single domain with domain-slot pairs predefined. In multi-domain DST, some of the previous works study the scalability of the model~\cite{wu2019transferable}, some aim to fully utilizing the dialogue history and context~\cite{shan2020contextual,chen2020parallel,quan2020modeling}, and some attempt to explore the relationship between different slots~\cite{hu2020sas,chen2020schema}. Nevertheless, existing approaches generally predict the dialogue state at every turn from scratch. The overwhelming majority of the slots in each turn should simply inherit the slot values from the previous turn. Therefore, the mechanism of treating slots equally in each turn not only is inefficient but also may lead to additional errors because of the redundant slot value generation.

To address this problem, we propose a DSS-DST which consists of the Dual Slot Selector based on the current turn dialogue, and the Slot Value Generator based on the dialogue history. At each turn, all slots are judged by the Dual Slot Selector first, and only the selected slots are permitted to enter the Slot Value Generator to update their slot value, while the other slots directly inherit the slot value from the previous turn. The Dual Slot Selector is a two-stage judging process. It consists of a Preliminary Selector and an Ultimate Selector, which jointly make a judgment for each slot according to the current turn dialogue. The intuition behind this design is that the Preliminary Selector makes a coarse judgment to exclude most of the irrelevant slots, and then the Ultimate Selector makes an intensive judgment for the slots selected by the Preliminary Selector and combines its confidence with the confidence of the Preliminary Selector to yield the final decision. Specifically, the Preliminary Selector briefly touches on the relationship of current turn dialogue utterances and each slot. Then the Ultimate Selector obtains a temporary slot value for each slot and calculates its reliability. The rationale for the Ultimate Selector is that if a slot value with high reliability can be obtained through the current turn dialogue, then the slot ought to be updated. Eventually, the selected slots enter the Slot Value Generator and a hybrid way of the extractive method and the classification-based method is utilized to generate a value according to the current dialogue utterances and dialogue history.

Our proposed DSS-DST achieves state-of-the-art joint accuracy on three of the most actively studied datasets: MultiWOZ 2.0~\cite{budzianowski2018multiwoz}, MultiWOZ 2.1~\cite{eric2019multiwoz}, and MultiWOZ 2.2~\cite{zang2020multiwoz} with joint accuracy of 56.93\%, 60.73\%, and 58.04\%. The results outperform the previous state-of-the-art by +2.54\%, +5.43\%, and +6.34\%, respectively. Furthermore, a series of subsequent ablation studies and analysis are conducted to demonstrate the effectiveness of the proposed method.

Our contributions in this paper are three folds:

\begin{itemize}
    \item We devise an effective DSS-DST which consists of the Dual Slot Selector based on the current turn dialogue and the Slot Value Generator based on the dialogue history to alleviate the redundant slot value generation.
	\item We propose two complementary conditions as the base of the judgment, which significantly improves the performance of the slot selection.
	\item Empirical results show that our model achieves state-of-the-art performance with significant improvements.
\end{itemize}

\section{Related Work}
Traditional statistical dialogue state tracking models combine semantics extracted by spoken language understanding modules to predict the current dialogue state~\cite{williams2007partially,thomson2010bayesian,wang2013simple,williams2014web} or to jointly learn speech understanding~\cite{henderson2014word,zilka2015incremental,wen2017network}.  With the recent development of deep learning and representation learning, most works about DST focus on encoding dialogue context with deep neural networks and predicting a value for each possible slot
~\cite{xu2018end,zhong2018global,ren2018towards,xie2018cost}. For multi-domain DST, slot-value pairs are extended to domain-slot-value pairs for the target~\cite{ramadan2018large,gao2019dialog,wu2019transferable,chen2020schema,hu2020sas,heck2020trippy,zhang2020find}. These models greatly improve the performance of DST, but the mechanism of treating slots equally is inefficient and may lead to additional errors. SOM-DST~\cite{kim2020efficient} considered the dialogue state as an explicit fixed-size memory and proposed a selectively overwriting mechanism. Nevertheless, it arguably has limitations because it lacks the explicit exploration of the relationship between slot selection and local dialogue information.

On the other hand, dialogue state tracking and machine reading comprehension (MRC) have similarities in many aspects~\cite{gao2020machine}. In MRC task, unanswerable questions are involved, some studies pay attention to this topic with straightforward solutions. \cite{liu2018stochastic} appended an empty word token to the context and added a simple classification layer to the reader. \cite{hu2019read+} used two types of auxiliary loss to predict plausible answers and the answerability of the question. \cite{zhang2020retrospective} proposed a retrospective reader that integrates both sketchy and intensive reading. \cite{zhang2020sg} proposed a verifier layer to context embedding weighted by start and end distribution over the context words representations concatenated to $[\mathrm{CLS}]$ token representation for BERT. The slot selection and the mechanism of local reliability verification in our work are inspired by the answerability prediction in machine reading comprehension.

\begin{figure*}[t]
\centering
\includegraphics[width=0.9\textwidth]{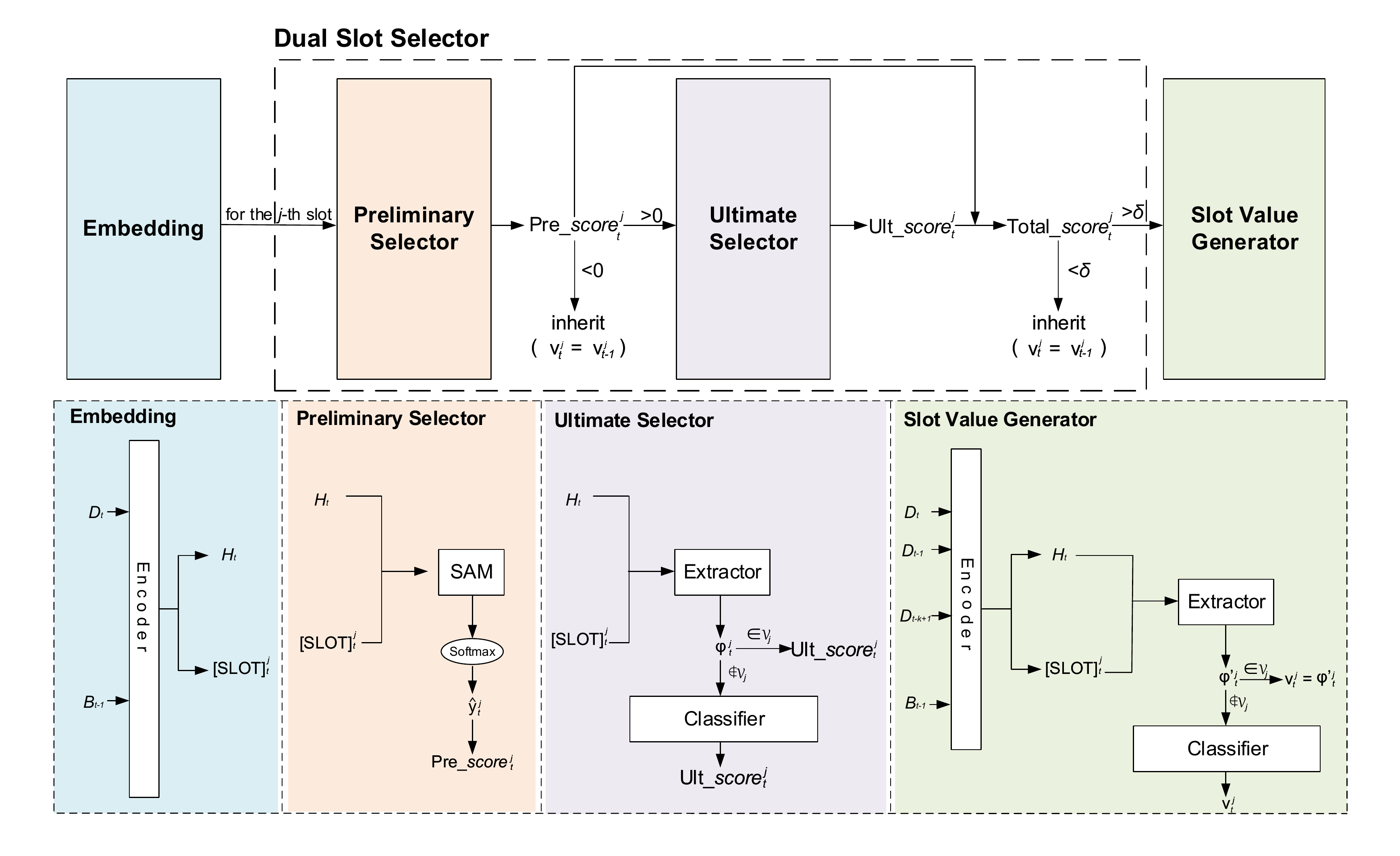}
\caption{The architecture of the proposed DSS-DST model. The upper part of the figure is the process between each module. The four blocks in the lower part of the figure are the internal structures of the modules with the same color above. At each turn, all slots are judged first, and the slots selected to be updated are permitted to enter the Slot Value Generator to update slot values, while the other slots directly inherit the slot values from the previous turn. The input utterances of the Slot Value Generator are the dialogues of the previous $k-1$ turns and the current turn, while the Dual Slot Selector only utilizes the current turn dialogue as the input utterances.}
\label{fig:framework}
\end{figure*}

\section{The Proposed Method}

Figure~\ref{fig:framework} illustrates the architecture of DSS-DST. DSS-DST consists of Embedding, Dual Slot Selector, and Slot Value Generator. In the task-oriented dialogue system, given a dialogue $Dial=\{(U_1,R_1);(U_2,R_2)\ldots;(U_T,R_T)\}$ of $T$ turns where $U_t$ represents user utterance and $R_t$ represents system response of turn $t$. We define the dialogue state at turn $t$ as $\mathcal{B}_t=\{(S^j,V_t^j)|1\le j\le J\}$,  where $S^j$ are the slots, $V_t^j$ are the corresponding slot values, and $J$ is the total number of such slots. Following \cite{lee2019sumbt}, we use the term “slot” to refer to the concatenation of a domain name and a slot name (e.g., “$restaurant-food$”).

\subsection{Embedding}

We employ the representation of the previous turn dialog state $B_{t-1}$ concatenated to the representation of the current turn dialogue $D_t$ as input:
\begin{equation}
    X_t=[\mathrm{CLS}]\oplus D_t\oplus B_{t-1}
\end{equation}
where $[\mathrm{CLS}]$ is a special token added in front of every turn input. Following SOM-DST~\cite{kim2020efficient}, we denote the representation of the dialogue at turn $t$ as $D_t=R_t\oplus;\oplus U_t\oplus[\mathrm{SEP}]$, where $R_t$ is the system response and $U_t$ is the user utterance. $;$ is a special token used to mark the boundary between $R_t$ and $U_t$, and $[\mathrm{SEP}]$ is a special token used to mark the end of a dialogue turn. The representation of the dialogue state at turn $t$ is $B_t=B_t^1\oplus\ldots\oplus B_t^J$, where $B_t^j=[\mathrm{SLOT}]^j\oplus S_j\oplus -\oplus V_t^j$ is the representation of the $j$-th slot-value pair. $-$ is a special token used to mark the boundary between a slot and a value. $[\mathrm{SLOT}]^j$ is a special token that represents the aggregation information of the $j$-th slot-value pair. We feed a pre-trained ALBERT~\cite{lan2019albert} encoder with the input $X_t$. Specifically, the input text is first tokenized into subword tokens. For each token, the input is the sum of the input tokens $X_t$ and the segment id embeddings. For the segment id, we use 0 for the tokens that belong to $B_{t-1}$ and 1 for the tokens that belong to $D_t$.

The output representation of the encoder is $O_t\in\mathbb{R}^{|X_t|\times d}$, and $h_t^{[\mathrm{CLS}]},h_t^{[\mathrm{SLOT}]^j}\in\mathbb{R}^d$ are the outputs that correspond to $[\mathrm{CLS}]$ and $[\mathrm{SLOT}]^j$, respectively. To obtain the representation of each dialogue and state, we split the $O_t$ into $H_t$ and $H_{t-1}^B$ as the output representations of the dialogue at turn $t$ and the dialogue state at turn $t-1$.

\subsection{Dual Slot Selector}

The Dual Slot Selector consists of a Preliminary Selector and an Ultimate Selector, which jointly make a judgment for each slot according to the current turn dialogue.

\paragraph{Slot-Aware Matching}
Here we first describe the Slot-Aware Matching (SAM) layer, which will be used as the subsequent components. The slot can be regarded as a special category of questions, so inspired by the previous success of explicit attention matching between passage and question in MRC~\cite{kadlec2016text,dhingra2017gated,wang2017gated,seo2016bidirectional}, we feed a representation $H$ and the output representation $h_t^{[\mathrm{SLOT}]^j}$ at turn $t$ to the Slot-Aware Matching layer by taking the slot presentation as the attention to the representation $H$:
\begin{equation}
    \mathrm{SAM}(H,j,t)=\mathrm{softmax}(H(h_t^{[\mathrm{SLOT}]^j})^\intercal)
\end{equation}
The output represents the correlation between each position of $H$ and the $j$-th slot at turn $t$.

\paragraph{Preliminary Selector}
The Preliminary Selector briefly touches on the relationship of current turn dialogue utterances and each slot to make an initial judgment. For the $j$-th slot $(1\le j\le J)$ at turn $t$, we feed its output representation $h_t^{[\mathrm{SLOT}]^j}$ and the dialogue representation $H_t$ to the SAM as follows:
\begin{equation}
    \bm{\alpha}_t^j\mathrm=\mathrm{SAM}(H_t,j,t)
\end{equation}
where $\bm{\alpha}_t^j\in\mathbb{R}^{N\times1}$ denotes the correlation between each position of the dialogue and the $j$-th slot at turn $t$. Then we get the aggregated dialogue representation $H_t^j\in\mathbb{R}^{N\times d}$ and passed it to a fully connected layer to get classification the $j$-th slot's logits $\hat{y}_t^j$ composed of selected ($logit\_\mathrm{sel}_t^i$) and fail ($logit\_\mathrm{fai}_t^j$) elements as follows:
\begin{gather}
    H_t^j,{_m}=\bm{\alpha}_t^j,_mH_t,_m,\ 0\le m<N \\
    \hat{y}_t^j=\mathrm{softmax}(\mathrm{FC}(H_t^j))
\end{gather}
We calculate the difference as the Preliminary Selector score for the $j$-th slot at turn $t$: $\mathrm{Pre}\_score_t^j=logit\_\mathrm{sel}_t^j-logit\_\mathrm{fai}_t^j$, and define the set of the slot indices as $U_{1,t}=\{j|\mathrm{Pre}\_score_t^j>0\}$, and its size as $J_{1,t}=|U_{1,t}|$. In the next paragraph, the slot in $U_{1,t}$ will be processed as the target object of the Ultimate Selector.

\paragraph{Ultimate Selector}
The Ultimate Selector will make the judgment on the slots in $U_{1,t}$. The mechanism of the Ultimate Selector is to obtain a temporary slot value for the slot and calculate its reliability through the dialogue at turn $t$ as its confidence for each slot. Specifically, for the $j$-th slot in $U_{1,t}$ ($1\le j\le J_{1,t}$), we first attempt to obtain the temporary slot value $\varphi_t^j$ using the extractive method:
We employ two different linear layers and feed $H_t$ as the input to obtain the representation $H\_\mathrm{s}_t$ and $H\_\mathrm{e}_t$ for predicting the start and end, respectively. Then we feed them to the SAM with the $j$-th slot to obtain the correlation representation $\bm{\alpha}\_\mathrm{s}_t^j$ and $\bm{\alpha}\_\mathrm{e}_t^j$ as follows:
\begin{gather}
    H\_\mathrm{s}_t=W_t^\mathrm{s}H_t \\
    H\_\mathrm{e}_t=W_t^\mathrm{e}H_t \\
    \bm{\alpha}\_\mathrm{s}_t^j=\mathrm{SAM}(H\_\mathrm{s}_t,j,t) \\
    \bm{\alpha}\_\mathrm{e}_t^j=\mathrm{SAM}(H\_\mathrm{e}_t,j,t)
\end{gather}
The position of the maximum value in $\bm{\alpha}\_\mathrm{s}_t^j$ and $\bm{\alpha}\_\mathrm{e}_t^j$ will be the start and end predictions of $\varphi_t^j$:
\begin{gather}
    \mathrm{ps}_t^j=\argmax_m(\bm{\alpha}\_\mathrm{s}_t^j,_m) \\
    \mathrm{pe}_t^j=\argmax_m(\bm{\alpha}\_\mathrm{e}_t^j,_m) \\
    \varphi_t^j=Dial_t[\mathrm{ps}_t^j:\mathrm{pe}_t^j]
\end{gather}
Here we define $\mathcal{V}_j$, the candidate value set of the $j$-th slot. If $\varphi_t^j$ belongs to $\mathcal{V}_j$, we calculate its proportion of all possible extracted temporary slot values and calculate the $\mathrm{Ult}\_score_t^j$ as the score of the $j$-th slot:
\begin{small}
\begin{gather}
    logit\_\mathrm{span}_t^j=\frac{\exp(\bm{\alpha}\_\mathrm{s}_t^j[\mathrm{ps}_t^j]+\bm{\alpha}\_\mathrm{e}_t^j[\mathrm{pe}_t^j])}{\sum\limits_{p_1=0}^{N-1}\sum\limits_{p_2=p_1+1}^{N-1}\exp(\bm{\alpha}\_\mathrm{s}_t^j[p_1]+\bm{\alpha}\_\mathrm{e}_t^j[p_2])} \label{eq:logit_span} \\
    logit\_\mathrm{null}_t^j=\frac{\exp(\bm{\alpha}\_\mathrm{s}_t^j[0]+\bm{\alpha}\_\mathrm{e}_t^j[0])}{\sum\limits_{p_1=0}^{N-1}\sum\limits_{p_2=p_1+1}^{N-1}\exp(\bm{\alpha}\_\mathrm{s}_t^j[p_1]+\bm{\alpha}\_\mathrm{e}_t^j[p_2])}
\end{gather}
\end{small}
\begin{equation}
    \mathrm{Ult}\_score_t^j=logit\_\mathrm{span}_t^j-logit\_\mathrm{null}_t^j
\end{equation}
If $\varphi_t^j$ does not belong to $\mathcal{V}_j$, we employ the classification-based method instead to select a temporary slot value from $\mathcal{V}_j$. Specifically, the dialogue representation $H_t^j$ is passed to a fully connected layer to get the distribution of $\mathcal{V}_j$. We choose the candidate slot value corresponding to the maximum value as the new temporary slot value $\varphi_t^j$, and calculate the distribution probability difference between $\varphi_t^j$ and ``$None$'' as the $\mathrm{Ult}\_score_t^j$:
\begin{gather}
    \bm{\alpha}\_\mathrm{c}_t^j=\mathrm{softmax}(\mathrm{FC}(H_t^j)) \\
    max\mathrm{c}=\argmax_m(\bm{\alpha}\_\mathrm{c}_t^j,_m) \\
    \mathrm{Ult}\_score_t^j=\bm{\alpha}\_\mathrm{c}_t^j[max\mathrm{c}]-\bm{\alpha}\_\mathrm{c}_t^j[0]
\end{gather}
We choose 0 as index because $\mathcal{V}_j[0]=``None"$.

\paragraph{Threshold-based decision}
Following previous studies~\cite{devlin2019bert,yang2019xlnet,liu2019roberta,lan2019albert}, we adopt the threshold-based decision to make the final judgment for each slot in $U_{1,t}$. The slot-selected threshold $\delta$ is set and determined in our model. The total score of the $j$-th slot is the combination of the predicted Preliminary Selector's score and the predicted Ultimate Selector's score:
\begin{equation}
    \mathrm{Total}\_score_t^j=\beta\mathrm{Pre}\_score_t^j+(1-\beta)\mathrm{Ult}\_score_t^j
\end{equation}
where $\beta$ is the weight. We define the set of the slot indices as $U_{2,t}=\{j|\mathrm{Total}\_score_t^j>\delta\}$, and its size as $J_{2,t}=|U_{2,t}|$. The slot in $U_{2,t}$ will enter the Slot Value Generator to update the slot value.

\subsection{Slot Value Generator}

After the judgment of the Dual Slot Selector, the slots in $U_{2,t}$ are the final selected slots. For each $j$-th slot in $U_{2,t}$, the Slot Value Generator generates a value for it. Conversely, the slots that are not in $U_{2,t}$ will inherit the slot value of the previous turn (i.e., $V_t^i=V_{t-1}^i,1\le i\le J-J_{2,t}$). For the sake of simplicity, we sketch the process as follows because this module utilizes the same hybrid way of the extractive method and the classification-based method as in the Ultimate Selector:
\begin{gather}
    X\_g_t=[\mathrm{CLS}]\oplus D_t\oplus \dots \oplus D_{t-k+1}\oplus B_{t-1} \\
    H\_g_t=Embedding(\mathrm{X}\_g_t) \\
    \mathrm{\varphi}\_g_t^j=Ext\_method(H\_g_t),1\leq j\leq J_{2,t} \\
    V_t^j=\mathrm{\varphi}\_\mathrm{g}_t^j\ ,\ \mathrm{\varphi}\_\mathrm{g}_t^j\in \mathcal{V}_j \\
    V_t^j=Cls\_method(H\_g_t)\ ,\ \mathrm{\varphi}\_\mathrm{g}_t^j\notin \mathcal{V}_j
\end{gather}

Significantly, the biggest difference between the Slot Value Generator and the Ultimate Selector is that the input utterances of the Slot Value Generator are the dialogues of the previous $k-1$ turns and the current turn, while the Ultimate Selector only utilizes the current turn dialogue as the input utterances.

\subsection{Optimization}

During training, we optimize both Dual Slot Selector and Slot Value Generator.

\paragraph{Preliminary Selector}
We use cross-entropy as a training objective:
\begin{equation}
    L_{\mathrm{pre},t}=-\frac{1}{J}\sum\limits_{j=1}^J[y_t^j\log \hat{y}_t^j+(1-y_t^j)\log (1-\hat{y}_t^i)]
\end{equation}
where $\hat{y}_t^j$ denotes the prediction and $y_t^j$ is the target indicating whether the slot is selected.

\paragraph{Ultimate Selector}
The training objectives of both extractive method and classification-based method are defined as cross-entropy loss:
\begin{gather}
    L_{\mathrm{ext},t}=-\frac{1}{J_{1,t}}\sum\limits_{j}^{J_{1,t}}\log(\mathrm{logit}\_p_t^j) \\
    L_{\mathrm{cls},t}=-\frac{1}{J_{1,t}}\sum\limits_{j}^{J_{1,t}}\sum\limits_{i}^{|\mathcal{V}_j|}y\_\mathrm{c}_{t,i}^j\log \bm{\alpha}\_\mathrm{c}_{t,i}^j
\end{gather}
where $\mathrm{logit}\_p_t^j$ is the target indicating the proportion of all possible extracted temporary slot values which is calculated according to the form of Equation~\ref{eq:logit_span}, and $y\_\mathrm{c}_{t,i}^j$ is the target indicating the probability of candidate values.

\paragraph{Slot Value Generator}
The training objective $L_{\mathrm{gen},t}$ of this module has the same form of training objective as in the Ultimate Selector.


\section{Experimental Setup}

\subsection{Datasets and Metrics}
We choose MultiWOZ 2.0~\cite{budzianowski2018multiwoz}, MultiWOZ 2.1~\cite{eric2019multiwoz}, and the latest MultiWOZ 2.2~\cite{zang2020multiwoz} as our training and evaluation datasets. These are the three largest publicly available multi-domain task-oriented dialogue datasets, including over 10,000 dialogues, 7 domains, and 35 domain-slot pairs. MultiWOZ 2.1 fixes the previously existing annotation errors. MultiWOZ 2.2 is the latest version of this dataset. It identifies and fixes the annotation errors of dialogue states on MultiWOZ2.1, solves the inconsistency of state updates and the problems of ontology, and redefines the dataset by dividing all slots into two types: non-categorical and categorical. In conclusion, it helps make a fair comparison between different models and will be crucial in the future research of this field.

Following TRADE~\cite{wu2019transferable}, we use five domains for training, validation, and testing, including \emph{restaurant}, \emph{train}, \emph{hotel}, \emph{taxi}, \emph{attraction}. These domains contain 30 slots (i.e., $J=30$). We use joint accuracy and slot accuracy as evaluation metrics. Joint accuracy refers to the accuracy of the dialogue state in each turn. Slot accuracy only considers individual slot-level accuracy.

\subsection{Baseline Models}
We compare the performance of DSS-DST with the following competitive baselines:

\noindent \textbf{DSTreader} formulates the problem of DST as an extractive QA task and extracts the value of the slots from the input as a span~\cite{gao2019dialog}. \textbf{TRADE} encodes the whole dialogue context and decodes the value for every slot using a copy-augmented decoder~\cite{wu2019transferable}. \textbf{NADST} uses a Transformer-based non-autoregressive decoder to generate the current turn dialogue state~\cite{le2019non}. \textbf{PIN} integrates an interactive encoder to jointly model the in-turn dependencies and cross-turn dependencies~\cite{chen2020parallel}. \textbf{DS-DST} uses two BERT-base encoders and takes a hybrid approach~\cite{zhang2020find}. \textbf{SAS} proposes a Dialogue State Tracker with Slot Attention and Slot Information Sharing to reduce redundant information’s interference~\cite{hu2020sas}. \textbf{SOM-DST} considers the dialogue state as an explicit fixed-size memory and proposes a selectively overwriting mechanism~\cite{kim2020efficient}. \textbf{DST-Picklist} performs matchings between candidate values and slot-context encoding by considering all slots as picklist-based slots~\cite{zhang2020find}. \textbf{SST} proposes a schema-guided multi-domain dialogue state tracker with graph attention networks~\cite{chen2020schema}. \textbf{TripPy} extracts all values from the dialog context by three copy mechanisms~\cite{heck2020trippy}.

\begin{table*}[t]
\centering
\begin{tabular}{l|cc|cc|cccc}
\hline
\multicolumn{1}{c|}{\multirow{2}{*}{\textbf{Model}}} & \multicolumn{2}{c|}{\textbf{MultiWOZ 2.0}}                                                                                   & \multicolumn{2}{c|}{\textbf{MultiWOZ 2.1}}                                                                                   & \multicolumn{4}{c}{\textbf{MultiWOZ 2.2}}                                                                                                                                                                                                                              \\ \cline{2-9} 
                                & \textbf{\begin{tabular}[c]{@{}c@{}}Joint\end{tabular}} & \textbf{\begin{tabular}[c]{@{}c@{}}Slot\end{tabular}} & \textbf{\begin{tabular}[c]{@{}c@{}}Joint\end{tabular}} & \textbf{\begin{tabular}[c]{@{}c@{}}Slot\end{tabular}} & \textbf{\begin{tabular}[c]{@{}c@{}}Joint\end{tabular}} & \textbf{\begin{tabular}[c]{@{}c@{}}Slot\end{tabular}} & \textbf{\begin{tabular}[c]{@{}c@{}}Cat-joint\end{tabular}} & \textbf{\begin{tabular}[c]{@{}c@{}}Noncat- \\joint\end{tabular}} \\ \hline
DSTreader                     & 39.41                                                         & -                                                            & 36.40                                                         & -                                                            & -                                                             & -                                                            & -                                                                 & -                                                                    \\
TRADE                         & 48.60                                                         & 96.92                                                        & 45.60                                                         & -                                                            & 45.40                                                         & -                                                            & 62.80                                                             & 66.60                                                                \\
NADST                         & 50.52                                                         & -                                                            & 49.04                                                         & -                                                            & -                                                             & -                                                            & -                                                                 & -                                                                    \\
PIN                           & 52.44                                                         & 97.28                                                        & 48.40                                                         & 97.02                                                        & -                                                             & -                                                            & -                                                                 & -                                                                    \\
DS-DST                        & -                                                             & -                                                            & 51.21                                                         & 97.35                                                        & 51.70                                                         & -                                                            & 70.60                                                             & 70.10                                                                \\
SAS                           & 51.03                                                         & 97.20                                                        & -                                                             & -                                                            & -                                                             & -                                                            & -                                                                 & -                                                                    \\
SOM-DST                       & 52.32                                                         & -                                                            & 53.68                                                         & -                                                            & -                                                             & -                                                            & -                                                                 & -                                                                    \\
DST-Picklist                  & 54.39                                                         & -                                                            & 53.30                                                         & 97.40                                                        & -                                                             & -                                                            & -                                                                 & -                                                                    \\
SST                           & 51.17                                                         & -                                                            & 55.23                                                         & -                                                            & -                                                             & -                                                            & -                                                                 & -                                                                    \\
TripPy                        & -                                                             & -                                                            & 55.30                                                         & -                                                            & -                                                             & -                                                            & -                                                                 & -                                                                    \\ \hline
DSS-DST                  & \begin{tabular}[c]{@{}c@{}}\textbf{56.93}\\ ($\pm$0.43) \end{tabular}                                                & \begin{tabular}[c]{@{}c@{}}\textbf{97.55}\\ ($\pm$0.05) \end{tabular}                                               & \begin{tabular}[c]{@{}c@{}}\textbf{60.73}\\ ($\pm$0.51) \end{tabular}                                                & \begin{tabular}[c]{@{}c@{}}\textbf{98.05}\\ ($\pm$0.06) \end{tabular}                                               & \begin{tabular}[c]{@{}c@{}}\textbf{58.04}\\ ($\pm$0.49) \end{tabular}                                                & \begin{tabular}[c]{@{}c@{}}\textbf{97.66}\\ ($\pm$0.06) \end{tabular}                                               & \begin{tabular}[c]{@{}c@{}}\textbf{76.32}\\ ($\pm$0.27) \end{tabular}                                                    & \begin{tabular}[c]{@{}c@{}}\textbf{73.39}\\ ($\pm$0.32) \end{tabular}                                                       \\ \hline
\end{tabular}
\caption{Joint accuracy (\%) and slot accuracy (\%) on the test sets of MultiWOZ 2.0, 2.1, and 2.2 vs. various approaches as reported in the literature. Cat-joint and noncat-joint denote joint accuracy on categorical and non-categorical slots, respectively.}
\label{table:table_1}
\end{table*}

\begin{table}[t]
\centering
\begin{tabular}{ll}
\hline
\begin{tabular}[c]{@{}c@{}}Pre-Trained\\ Language Model\end{tabular} & MultiWOZ 2.1   \\ \hline
Our Model                                                              & \textbf{60.73} \\
BERT (large)                                                         & 60.11 (-0.62)  \\
ALBERT (base)                                                        & 59.98 (-0.75)  \\
BERT (base)                                                          & 59.35 (-1.38)  \\ \hline
\end{tabular}
\caption{The ablation study of the DSS-DST on the MultiWOZ 2.1 dataset with joint accuracy (\%).}
\label{table:table_2}
\end{table}

\begin{table}[t]
\centering
\begin{tabular}{ll}
\hline
\multicolumn{1}{c}{\begin{tabular}[c]{@{}c@{}} Model\end{tabular}} & MultiWOZ 2.1   \\ \hline
Our Model                                                          & \textbf{60.73} \\
-Ultimate Selector                                                 & 58.82 (-1.91)  \\
-Preliminary Selector                                              & 52.22 (-8.51)  \\
-above two                                                         & 40.69 (-20.04)  \\ \hline
\end{tabular}
\caption{The ablation study of the DSS-DST on the MultiWOZ 2.1 dataset with joint accuracy (\%).}
\label{table:table_3}
\end{table}

\subsection{Training}
We employ a pre-trained ALBERT-large-uncased model~\cite{lan2019albert} for the encoder of each part. The hidden size of the encoder $d$ is 1024. We use AdamW optimizer~\cite{loshchilov2018fixing} and set the warmup proportion to 0.01 and L2 weight decay of 0.01. We set the peak learning rate to 0.03 for the Preliminary Selector and 0.0001 for the Ultimate Selector and the Slot Value Generator, respectively. The max-gradient normalization is utilized and the threshold of gradient clipping is set to 0.1. We use a batch size of 8 and set the dropout~\cite{srivastava2014dropout} rate to 0.1. In addition, we utilize word dropout~\cite{bowman2016generating} by randomly replacing the input tokens with the special [UNK] token with the probability of 0.1. The max sequence length for all inputs is fixed to 256.

We train the Preliminary Selector for 10 epochs and train the Ultimate Selector and the Slot Value Generator for 30 epochs. During training the Slot Value Generator, we use the ground truth selected slots instead of the predicted ones. We set $k$ to 2, $\beta$ to 0.55, and $\delta$ to 0. For all experiments, we report the mean joint accuracy over 10 different random seeds to reduce statistical errors.

\begin{table}[t]
\centering
\begin{tabular}{ll}
\hline
\multicolumn{1}{c}{\begin{tabular}[c]{@{}c@{}} Model\end{tabular}} & MultiWOZ 2.1   \\ \hline
Our Model                                                          & \textbf{60.73} \\
Dialogue History${}^\dagger$                                                   & 58.36 (-2.37)  \\ \hline
\end{tabular}
\caption{The ablation study of the DSS-DST on the MultiWOZ 2.1 dataset with joint accuracy (\%). $\dagger$ means attaching the dialogue of the previous turn to the current turn dialogue as the input of the Dual Slot Selector.}
\label{table:table_4}
\end{table}

\begin{table}[t]
\centering
\begin{tabular}{lc}
\hline
\multicolumn{1}{c}{\begin{tabular}[c]{@{}c@{}} $k$\end{tabular}} & MultiWOZ 2.1   \\ \hline
1                                                          & 53.96\\
2 (Our Model)                                              & \textbf{60.73}  \\
3                                                          & 59.34  \\\hline
\end{tabular}
\caption{The joint accuracy (\%) of different $k$ on MultiWOZ 2.1 dataset. The $k$ represents the dialogue history of the previous $k-1$ turns.}
\label{table:table_5}
\end{table}

\begin{table}[t]
\centering
\begin{tabular}{cccc}
\hline
\multicolumn{2}{c}{Our Model}                                              & \multicolumn{2}{c}{SOM-DST}                                             \\ \hline
\begin{tabular}[c]{@{}c@{}}Operation\end{tabular} & F1             & \begin{tabular}[c]{@{}c@{}}Operation\end{tabular} & F1    \\ \hline
\textit{inherit}                                                    & \textbf{99.71} & CARRYOVER                                                       & 98.66 \\ \hline
\textit{update}                                                     & \textbf{90.65} & UPDATE                                                          & 80.10  \\
                                                                    &                & DELETE                                                          & 32.51 \\
                                                                    &                & DONTCARE                                                        & 2.86  \\ \hline
\end{tabular}
\caption{Statistics of the state operations and the corresponding F1 scores of our model and SOM-DST in the test set of MultiWOZ 2.1.}
\label{table:table_6}
\end{table}

\begin{table}[t]
\centering
\begin{tabular}{cc}
\hline
\multicolumn{2}{c}{MultiWOZ 2.2} \\ \hline
Domain     & Joint Accuracy (\%) \\ \hline
Attraction & 79.88               \\
Hotel      & 62.47               \\
Restaurant & 75.79               \\
Taxi       & 54.84               \\
Train      & 76.25               \\ \hline
\end{tabular}
\caption{Domain-specific results on the test set of MultiWOZ 2.2. We are the first to list Domain-specific results on the test set of MultiWOZ 2.2 to the best of our knowledge.}
\label{table:table_7}
\end{table}

\begin{table}[t]
\centering
\begin{tabular}{lcc}
\hline
\multicolumn{1}{c}{\multirow{2}{*}{Model}} & \multicolumn{2}{c}{MultiWOZ 2.2}                          \\ \cline{2-3} 
\multicolumn{1}{c}{}                       & \multicolumn{1}{c}{Joint} & \multicolumn{1}{c}{Cat-joint} \\ \hline
Our Model                                   & \textbf{58.04}                     & \textbf{76.32}                          \\
-Extractive Method                          & 50.01             & 66.15 \\
\hline
\end{tabular}
\caption{The ablation study of the DSS-DST on the MultiWOZ 2.2 dataset with joint accuracy (\%) and joint accuracy on categorical slots.}
\label{table:table_8}
\end{table}

\section{Experimental Results}

\subsection{Main Results}
Table~\ref{table:table_1} shows the joint accuracy and the slot accuracy of our model and other baselines on the test sets of MultiWOZ 2.0, 2.1, and 2.2. As shown in the table, our model achieves state-of-the-art performance on three datasets with joint accuracy of 56.93\%, 60.73\%, and 58.04\%, which has a significant improvement over the previous best joint accuracy. Particularly, the joint accuracy on MultiWOZ 2.1 beyond 60\%. Despite the sparsity of experimental result on MultiWOZ 2.2, our model still leads by a large margin in the existing public models. Similar to~\cite{kim2020efficient}, our model achieves higher joint accuracy on MultiWOZ 2.1 than that on MultiWOZ 2.0. For MultiWOZ 2.2, the joint accuracy of categorical slots is higher than that of non-categorical slots. This is because we utilize the hybrid way of the extractive method and the classification-based method to treat categorical slots. However, we can only utilize the extractive method for non-categorical slots since they have no ontology (i.e., candidate value set).

\subsection{Ablation Study}
\paragraph{Pre-trained Language Model}
For a fair comparison, we employ different pre-trained language models with different scales as encoders for training and testing on MultiWOZ 2.1 dataset. As shown in Table~\ref{table:table_2}, the joint accuracy of other implemented ALBERT and BERT encoders decreases in varying degrees. In particular, the joint accuracy of BERT-base-uncased decreased by 1.38\%, but still outperformed the previous state-of-the-art performance on MultiWOZ 2.1. The result demonstrates the effectiveness of DSS-DST.

\paragraph{Separate Slot Selector}
To explore the effectiveness of the Preliminary Selector and Ultimate Selector respectively, we conduct an ablation study of the two slot selectors on MultiWOZ 2.1. As shown in Table~\ref{table:table_3}, we observe that the performance of the separate Preliminary Selector is better than that of the separate Ultimate Selector. This is presumably because the Preliminary Selector is the head of the Dual Slot Selector, it is stable when it handles all slots. Nevertheless, the input of the Ultimate Selector is the slots selected by the Preliminary Selector, and its function is to make a refined judgment. Therefore, it will be more vulnerable when handling all the slots independently. In addition, when the two selectors are removed, the performance drops drastically. This demonstrates that the slot selection is integral before slot value generation.

\paragraph{Dialogue History for the Dual Slot Selector}
As aforementioned, we consider that the slot selection only depends on the current turn dialogue. In order to verify it, we attach the dialogue of the previous turn to the current turn dialogue as the input of the Dual Slot Selector. We observe in Table~\ref{table:table_4} that the joint accuracy decreases by 2.37\%, which implies the redundant information of dialogue history confuse the slot selection in the current turn.

\paragraph{Dialogue History for the Slot Value Generator}
We try the number from one to three for the $k$ to observe the influence of the selected dialogue history on the Slot Value Generator. As shown in Table~\ref{table:table_5}, the model achieves better performance on MultiWOZ 2.1 when $k=2,3$ than that of $k=1$. Furtherly, the performance of $k=2$ is better than that of $k=3$. We conjecture that the dialogue history far away from the current turn is little helpful because the relevance between two sentences in dialogue is strongly related to their positions.

The above ablation studies show that dialogue history confuses the Dual Slot Selector, but it plays a crucial role in the Slot Value Generator. This demonstrates that there are fundamental differences between the two processes, and confirms the necessity of dividing DST into these two sub-tasks.

\section{Analysis}

\subsection{Comparative Analysis of Slot Selector}
We analyze the performance of the Dual Slot Selector and compare it with other previous work in MultiWOZ 2.1. Here we choose the SOM-DST and list the state operations and the corresponding F1 scores as a comparison. The SOM-DST sets four state operations (i.e., CARRYOVER, DELETE, DONTCARE, UPDATE), while our model classifies the slots into two classes (i.e., $inherit$ and $update$). It means that DELETE, DONTCARE, and UPDATE in SOM-DST all correspond to $update$ in our model. As shown in Table~\ref{table:table_6}, our model still achieves superior performance when dealing with $update$ slots, which contain DONTCARE, DELETE, and other difficult cases.

\subsection{Domains and Ontology}
Table~\ref{table:table_7} shows the domain-specific results of our model on the latest MultiWOZ 2.2 dataset. We can observe that the performance of our model in $taxi$ domain is lower than that of the other four domains. We investigate the dataset and find that all the slots in $taxi$ domain are non-categorical slots. This indicates the reason that we can only utilize the extractive method for non-categorical slots since they have no ontology. Furthermore, we test the performance of using the separate classification-based method for categorical slots. As illustrated in Table~\ref{table:table_8}, the joint accuracy of our model and categorical slots decreased by 8.03\% and 10.17\%, respectively.

\section{Conclusion}
We introduce an effective two-stage DSS-DST which consists of the Dual Slot Selector based on the current turn dialogue, and the Slot Value Generator based on the dialogue history. The Dual Slot Selector determines each slot whether to update or to inherit based on the two conditions. The Slot Value Generator employs a hybrid method to generate new values for the slots selected to be updated according to the dialogue history. Our model achieves state-of-the-art performance of 56.93\%, 60.73\%, and 58.04\% joint accuracy with significant improvements (+2.54\%, +5.43\%, and +6.34\%) over previous best results on MultiWOZ 2.0, MultiWOZ 2.1, and MultiWOZ 2.2 datasets, respectively. The mechanism of a hybrid method is a promising research direction and we will exploit a more comprehensive and efficient hybrid method for slot value generation in the future.

\section*{Acknowledgements}
This work was supported by the National key research and development project (2017YFB1400603) and the Foundation for Innovative Research Groups of the National Natural Science Foundation of China (Grant No. 61921003). We thank the anonymous reviewers for their insightful comments.


\section*{Ethical Considerations}
The claims in this paper match the experimental results. The model utilizes the hybrid method for slot value generation, so it is universal and scalable to unseen domains, slots, and values. The experimental results can be expected to generalize.

\bibliographystyle{acl_natbib}
\bibliography{anthology,acl2021}

\clearpage

\appendix
\section*{Appendices}
\renewcommand{\thesubsection}{\Alph{subsection}}

\subsection{Accuracy per Slot on MultiWOZ 2.2 Testset}

\begin{table}[h]
\centering
\begin{tabular}{lr}
\hline
Domain-Slot            & Our Model \\ \hline
attraction-area        & 97.95     \\
attraction-name        & 93.38     \\
attraction-type        & 97.37     \\
hotel-area             & 97.29     \\
hotel-book day         & 100       \\
hotel-book people      & 100       \\
hotel-book stay        & 100       \\
hotel-internet         & 94.94     \\
hotel-name             & 95.29     \\
hotel-parking          & 95.26     \\
hotel-price range      & 97.67     \\
hotel-stars            & 97.98     \\
hotel-type             & 93.24     \\
restaurant-area        & 97.34     \\
restaurant-book day    & 100       \\
restaurant-book people & 100       \\
restaurant-book time   & 100       \\
restaurant-food        & 96.76     \\
restaurant-name        & 94.26     \\
restaurant-price range & 97.88     \\
taxi-arrive by         & 98.68     \\
taxi-departure         & 97.24     \\
taxi-destination       & 97.05     \\
taxi-leave at          & 99.25     \\
train-arrive by        & 96.63     \\
train-book people      & 100       \\
train-day              & 99.59     \\
train-departure        & 98.32     \\
train-destination      & 98.48     \\
train-leave at         & 94.14     \\ \hline
\end{tabular}
\caption{The detailed results of accuracy (\%) per slot on MultiWOZ 2.2 test set. We sort them according to their domains.}
\label{table:accuracy_per_slot}
\end{table}

\clearpage

\onecolumn
\subsection{Data Statistics}

\begin{table*}[h]
\centering
\begin{tabular}{cccccccc}
\hline
           &                                                                                                                                                   & \multicolumn{3}{c}{Dialogues} & \multicolumn{3}{c}{Turns} \\ \hline
Domain     & Slots                                                                                                                                             & Train    & Valid    & Test    & Train   & Valid  & Test   \\ \hline
Hotel      & \begin{tabular}[c]{@{}c@{}}price range,\\ type,\\ parking,\\ book stay,\\ book day,\\ book people,\\ area, stars,\\ internet,\\ name\end{tabular} & 3,381    & 416      & 394     & 14,793  & 1,781  & 1,756  \\ \hline
Attraction & \begin{tabular}[c]{@{}c@{}}area, name,\\ type\end{tabular}                                                                                        & 2,717    & 401      & 395     & 8,073   & 1,220  & 1,256  \\ \hline
Restaurant & \begin{tabular}[c]{@{}c@{}}food, price\\ range, area,\\ name, book\\ time, book\\ day, book\\ people\end{tabular}                                 & 3,813    & 438      & 437     & 15,367  & 1,708  & 1,726  \\ \hline
Taxi       & \begin{tabular}[c]{@{}c@{}}leave at,\\ destination,\\ departure,\\ arrive by\end{tabular}                                                         & 1,654    & 207      & 195     & 4,618   & 690    & 654    \\ \hline
Train      & \begin{tabular}[c]{@{}c@{}}destination,\\ day,\\ departure,\\ arrive by,\\ book people,\\ leave at\end{tabular}                                   & 3,103    & 484      & 494     & 12,133  & 1,972  & 1,976  \\ \hline
\end{tabular}
\caption{Data statistics of MultiWOZ 2.1.}
\label{table:data_statistics}
\end{table*}

\end{document}